\documentclass{article}
\usepackage[preprint]{neurips_2026}
\usepackage{amsmath,amssymb,amsfonts}
\usepackage{algorithmic}
\usepackage{graphicx}
\usepackage{subcaption}
\usepackage{booktabs, multirow, makecell}
\usepackage{textcomp}
\usepackage{xcolor}
\def\BibTeX{{\rm B\kern-.05em{\sc i\kern-.025em b}\kern-.08em
    T\kern-.1667em\lower.7ex\hbox{E}\kern-.125emX}}

\usepackage[utf8]{inputenc} % allow utf-8 input
\usepackage[T1]{fontenc}    % use 8-bit T1 fonts
\usepackage{hyperref}       % hyperlinks
\usepackage{url}            % simple URL typesetting
\usepackage{booktabs}       % professional-quality tables
\usepackage{amsfonts}       % blackboard math symbols
\usepackage{nicefrac}       % compact symbols for 1/2, etc.
\usepackage{microtype}      % microtypography
\usepackage[ruled,vlined]{algorithm2e} % for algorithms
\usepackage{setspace} % used to stretch out linespacing in algorithm2e
\usepackage{bm}
\usepackage{lipsum}
\usepackage{enumitem}
\usepackage{tcolorbox}
\tcbuselibrary{skins}
\usepackage{threeparttable}
\usepackage{bbm}
\usepackage{setspace}
\usepackage{multicol}
\usepackage{tabularx}
\usepackage{array}
\usepackage{booktabs}
\usepackage{cases}
\usepackage[skip=1pt, belowskip=0pt]{caption}
\usepackage{subcaption} % for subfigures
\usepackage{amsthm}
\usepackage{arydshln}
\usepackage{mathtools}
\usepackage{comment} 
\usepackage[normalem]{ulem} % for strikeout
\usepackage{float}
\definecolor{myblue}{rgb}{0.2,0.2,0.9}
\definecolor{mygreen}{rgb}{0.0328, 0.4758, 0.0539} 
\definecolor{myred}{rgb}{0.7, 0.0328, 0.0539}
\definecolor{mypurple}{rgb}{0.7, 0.0328, 0.7539}
\definecolor{myyellow}{rgb}{0.99, 0.7, 0.1}
\definecolor{myblue}{rgb}{0.3328, 0.3539, 0.7758}
\definecolor{myblue2}{rgb}{0.0328, 0.0539, 0.4758}
\definecolor{mygreen2}{rgb}{ 0.0328 0.4758 0.0539} 
\definecolor{mygreen3}{rgb}{ 0.0328 0.1758 0.0539} 
\definecolor{myred}{rgb}{0.4758, 0.0328, 0.0539}
\definecolor{myred2}{rgb}{0.75, 0.0328, 0.0539}

\newcommand{\ubf}{\bm{u}}
\newcommand{\xbf}{\bm{x}}

\newcommand{\ybf}{\bm{y}}
\newcommand{\Ibf}{\bm{I}}

\newcommand{\Abf}{\bm{A}}
\newcommand{\Bbf}{\bm{B}}
\newcommand{\Cbf}{\bm{C}}

\newcommand{\Hneurons}{h}

\newcommand{\Nstate}{n}

\newcommand{\Wbf}{\bm{W}}

\newcommand{\cin}{c_{in}}
\newcommand{\cout}{c_{out}}

\newcommand{\thetabf}{\boldsymbol{\theta}}
\newcommand{\phibf}{\boldsymbol{\phi}}
\newcommand{\Dcal}{\mathcal{D}}
\newcommand{\psibf}{\boldsymbol{\psi}}

\newcommand{\Umem}[1]{x[#1]}

\newcommand{\Sil}[1]{y[#1]} % {y_{out}[#1]}

\newcommand{\AbfR}[1]{\bm{A}_{r_{#1}}} 
\newcommand{\BbfR}[1]{\bm{B}_{r_{#1}}}

\newcommand{\DeltaRatio}{\rho}
\newcommand{\ctR}[1]{T_{r_{#1}}} % continous time step high resolution

\newcommand{\Tscentral}{T_c}
\newcommand{\Tsnode}{T_{\ell}}

\newcommand{\normal}{standard-}
\newcommand{\Normal}{Standard-}

\theoremstyle{remark}

\title{Federated Learning of Spiking Neural Networks under Heterogeneous Temporal Resolutions}
\author{%
  Sanja~Karilanova 
  \quad Subhrakanti~Dey
  \quad Ayça~Özçelikkale \\
  Department of Electrical Engineering, Uppsala University, Sweden \\
  \texttt{\{Sanja.Karilanova, Subhrakanti.Dey, Ayca.Ozcelikkale\}@angstrom.uu.se} \\
}

\begin{document}
\maketitle
\begin{abstract}
Spiking neural networks (SNNs) are biologically inspired energy-efficient models that use sparse binary spike-based communication between neurons, making them attractive for resource-constrained edge devices. Federated learning enables such devices to train collaboratively without sharing raw data. In time-series applications, edge devices often collect data at different time resolutions due to hardware and energy constraints. This temporal heterogeneity poses a fundamental challenge for federated learning: parameters learned at one temporal resolution do not necessarily transfer directly to another, which might result in the naive federated averaging being ineffective.
Targeting SNNs and, more broadly, deep networks with stateful neurons, we propose a federated learning framework that addresses this temporal resolution mismatch.  
We investigate how neuron parameters learned from data at different temporal resolutions and model aggregation should be integrated.
We evaluate the proposed framework across two SNN-native benchmark datasets (SHD and DVS-Gesture) under a range of resolution heterogeneity scenarios. Our results show that the proposed adaptation methods can substantially recover accuracy lost due to temporal mismatch, hence enabling each client to train at their local temporal resolution while remaining compatible with the global model. 
\end{abstract}
\section{Introduction}
\label{sec:Introduction}

% FL 
Federated learning (FL) is a distributed machine learning paradigm where multiple entities (clients) collaboratively train a model under the coordination of a central server while keeping their raw data local. Instead of sharing data, clients transmit local model parameters that are aggregated to improve the global model \cite{10.1561/2200000083}. This framework enables learning from decentralized data while preserving privacy and reducing communication of sensitive information, and has been widely considered for edge and distributed sensing applications \cite{9153560}.

% Neuromorphic
Many FL deployments operate on edge devices with strict constraints on energy consumption, memory, and computational resources. Conventional signal processing and machine learning methods are typically implemented on \textit{von Neumann} architectures, which often incur significant energy costs \cite{sudhakar_data_2023}. Neuromorphic computing, inspired by principles of biological neural systems, has emerged as a promising alternative for energy-efficient data processing \cite{Rajendran_2019, davies_advancing_2021, yik2024neurobench}. In these systems, information is represented using spikes (impulses), typically encoded as binary events over time, enabling low-latency and energy-efficient computation suitable for edge devices. The most popular model used which satisfies these properties is Spiking Neural Networks (SNNs). In SNNs stateful neurons arrange in neural network architecture communicate using spikes. 

\begin{figure}
    \centering
    \begin{subfigure}[b]{0.368\linewidth}
        \centering
        \includegraphics[width=\linewidth]{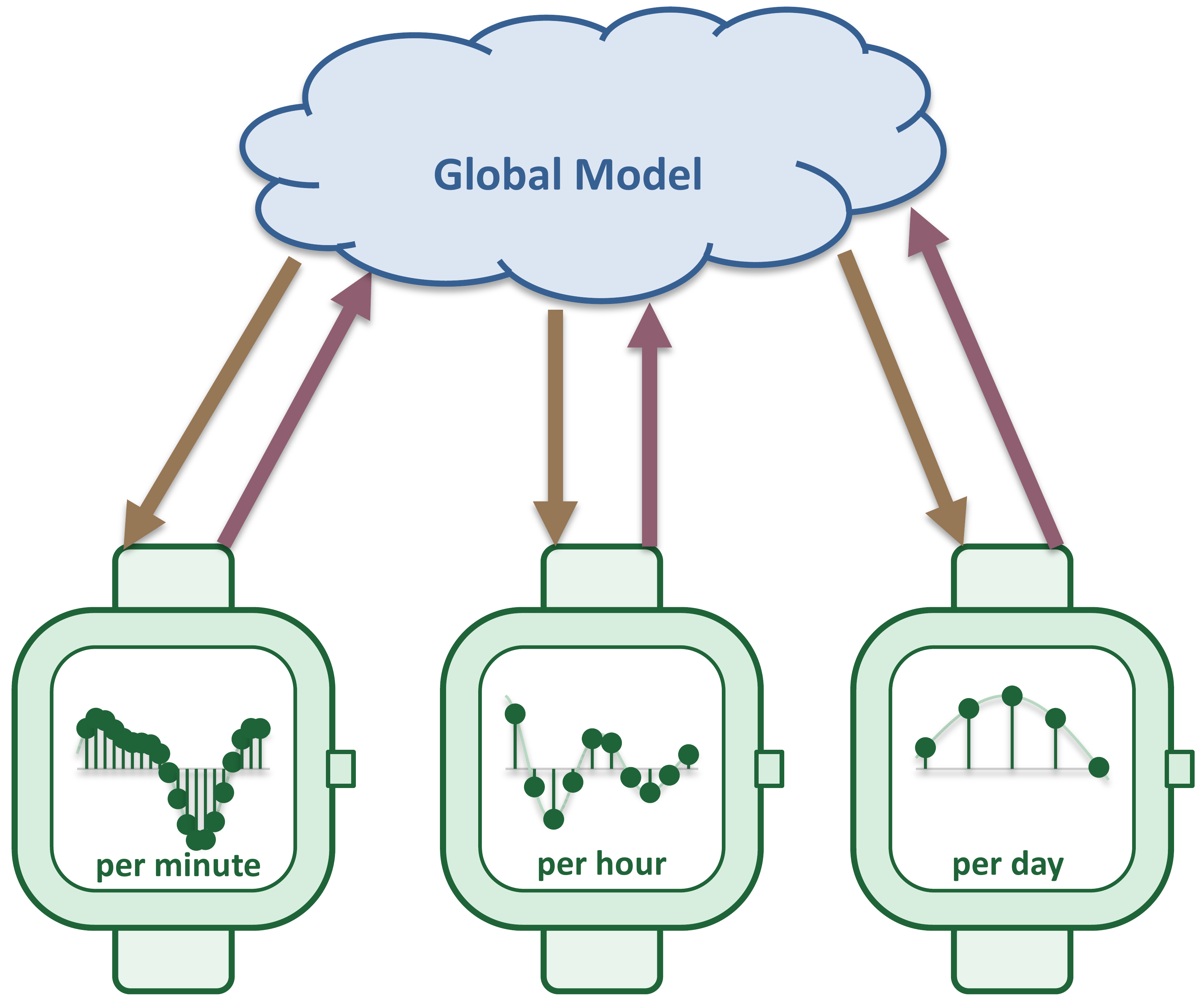}
        \caption{High level federated learning set-up}
        \label{fig:FL_set_up:high_level}
    \end{subfigure}
    \hfill
    \begin{subfigure}[b]{0.489\linewidth}
        \centering
        \includegraphics[width=\linewidth]{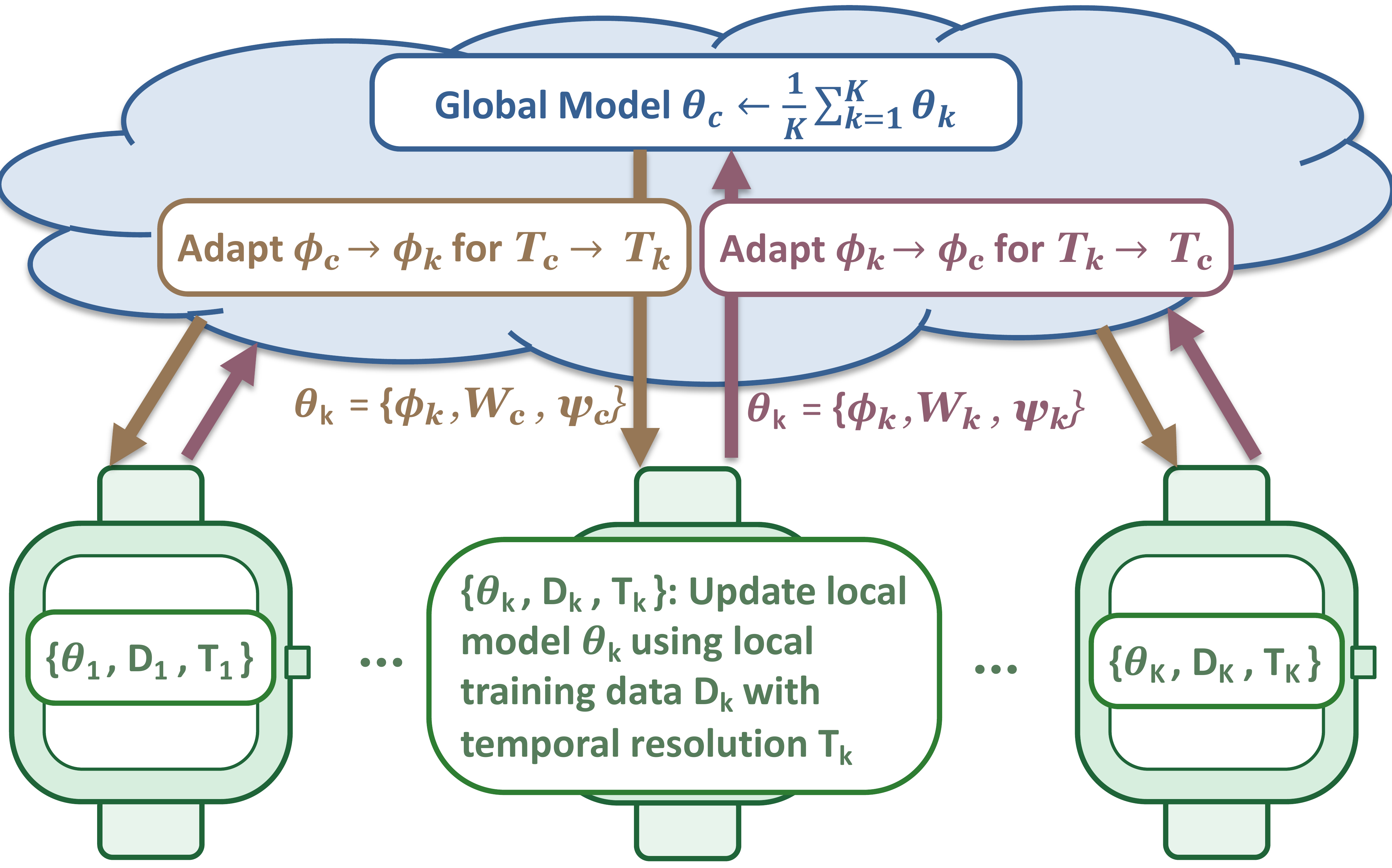}
        \caption{Overview of the proposed framework at one device}
        \label{fig:FL_set_up:detailed}
    \end{subfigure}
    \caption{Overview of the federated learning setup, illustrated with a smartwatch example.}
    \label{fig:FL_set_up}
\end{figure}

% Temporal resolution
In conventional data acquisition systems, \textit{temporal resolution} is determined by the sampling interval between successive observations, such as the frame rate of a video or the sampling rate of an audio signal. In practice, however, different devices often collect data at different temporal resolutions due to hardware limitations, energy constraints, memory capacity, or communication bandwidth restrictions \cite{Dieter_2005, ur2016BigDataCompress, chen2021distributed, park2021communication} (see Figure \ref{fig:FL_set_up:high_level} for a visualization example). Such discrepancies in temporal resolution can affect the performance of the global model \cite{Zubic_2024_CVPR, caccavella2023lowpower, karilanova2025zeroshottemporalresolutiondomain}. 

% Our contribution 
In this paper, we propose a federated learning framework for SNNs, and more broadly, stateful deep networks,  that explicitly accounts for the difference in temporal resolutions of the clients' data and the resulting temporal misalignment in the clients' trained networks. Our  \emph{Federated Learning with Temporal Resolution Adaptation~(FedTA)} framework allows each node to locally update the shared model using data at its own temporal resolution while maintaining compatibility with the global model for aggregation and evaluation.

\section{Related Work}

\paragraph{Federated Learning}

% Original version
Federated learning  commonly relies on parameter aggregation methods such as Federated Averaging (FedAvg) \cite{mcmahan2017communication}, in which clients perform local training and the server aggregates their local model parameters into a global model. FedAvg is one of the standard baselines for FL algorithms. Heterogeneity between clients during federated learning has been extensively studied in terms of data distribution \cite{10.5555/3666122.3666222, huang2026softconsensual, NEURIPS2023_32c2f3e0,10492865}, particularly class imbalance \cite{10492865}. 
% % % %
Nevertheless, to the best of our knowledge, existing studies do not address temporal resolution mismatch between clients and/or between center and clients. 

\paragraph{Deep State Space Models}
Deep state-space models (SSMs) have recently gained significant attention due to their strong performance in modeling sequential and temporal data \cite{gu2022efficientlymodelinglongsequences, gu2022parameterization, smith2023simplified}. Several works have studied how SSMs behave under varying sampling rates: the HiPPO-LegS parameterization \cite{gu2020hippo} targets timescale-invariant representations, \cite{Schone_EventbyEvent} adjusts the effective timestep for event-driven inputs, and \cite{Zubic_2024_CVPR} proposes resolution-aware training for sampling-robust SSMs on vision tasks. None of these works consider a federated setting. 
In parallel, emerging works combine SSM architectures with spiking neural representations, enabling efficient temporal processing while maintaining compatibility with neuromorphic hardware \cite{Zhang2026, SSM_Loihi, stan2024learning, Karilanova_NICE_MIMO}.

\paragraph{Spiking Neural Networks}
% TODO More SNN works/sucsess
In the context of SNNs, FedAvg has been explored but focuses primarily on the class imbalance between clients \cite{venkatesha2021federated, WANG2023126686, Liu_2022}, without considering differences in temporal resolution between clients. Separately, several works have addressed changes in temporal resolution in SNNs outside of federated settings \cite{karilanova2025zeroshottemporalresolutiondomain} \cite{caccavella2023lowpower} \cite{he2020comparing}. 
Leveraging the theoretical foundations in \cite{karilanova2025zeroshottemporalresolutiondomain},
we propose a federated learning approach that enables temporal alignment between clients with heterogeneous temporal resolution.
Federated learning tailored for deep SSMs and SNNs remains largely unexplored, and to the best of our knowledge, no prior work addresses heterogeneous temporal resolutions across clients. Our work addresses this gap.

\section{Problem Statement}
\label{sec:problem_statement}

We consider a federated learning setting with $K$ clients, where client $k$ collects time-series data using a client-specific sampling interval, i.e. temporal resolution, $T_k$. Every client trains a local deep network, and a central server aggregates the local models and sends the updated model to the clients back. Local models are naturally tied to temporal resolution: a model trained in $T_k$ may not generalize directly to data sampled in $T_j \neq T_k$. Naive parameter averaging therefore conflates models operating at different temporal resolutions, which can degrade the global model. This article focuses on the question: {\it{How should models, which are trained using data with different temporal resolutions, be adapted for model averaging under federated learning?}}

\begin{figure}
    \centering
    \includegraphics[width=1\linewidth]{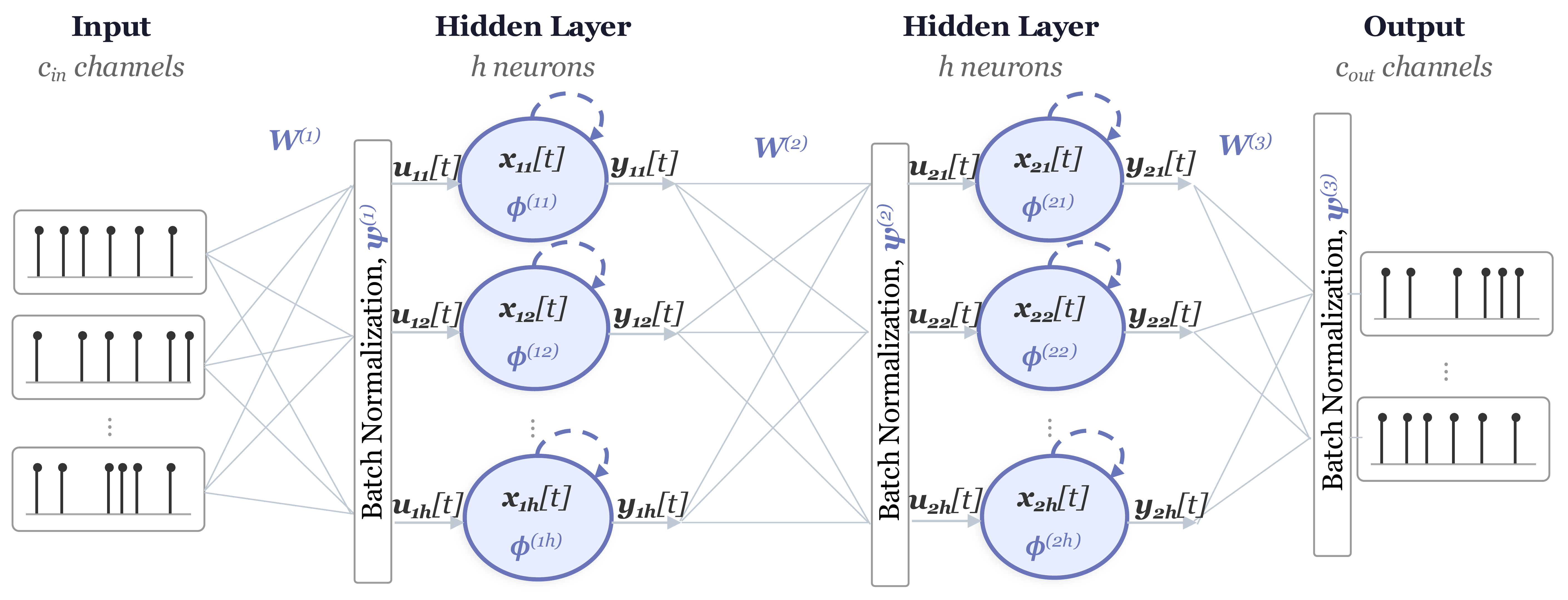}
    \caption{Illustrative Network Architecture}
    \label{fig:network_architecture}
\end{figure}

We study this problem under deep neural networks constructed by stacking hidden stateful layers, each consisting of $h$ stateful neurons with dynamical parameters $\phibf =\{ \phi^{(\ell i)} \}$, where $\ell$ is layer index and $i$ is neuron index in the layer connected through dense synaptic weights $\Wbf=\{ \Wbf^{(\ell)} \}$ and followed by a normalization layer parameterized by $\psibf=\{ \psibf^{(\ell )} \}$ (see Figure~\ref{fig:network_architecture}). We denote the set of full model parameters  as $\thetabf = \{ \phibf, \Wbf, \psibf\}$. Each neuron maintains an internal state across time following a possibly nonlinear discrete-time state-space model (SSM) of the form
\begin{subequations}
\label{eqn:SSM:nonlinear}
\begin{align}
    \xbf[t+1] &= f_{\phi}(\xbf[t],\, \ubf[t]), \\
    \ybf[t]   &= h_{\phi}(\xbf[t]),
\end{align}
\end{subequations}
where $\ubf[t]$ is the input to the neuron, $\xbf[t]$ the latent state of the neuron, and $\ybf[t]$ the output at time $t$. For clarity, we omit the layer and neuron indices throughout. Here, $\phi$  represent the trainable parameters of the neuron family which represents the combined list of parameters of $f_{\phi}(\cdot)$ and $h_{\phi}(\cdot)$. 
The specific parametric forms of $f_{\phi}(\cdot)$ and $h_{\phi}(\cdot)$ depend on the neuron model (See Section \ref{sec:SSM-based-NN} ). 
Under a federated averaging setting, we consider the following main question: 
{\it{How should the neuron parameters $\phi$, learned  using data with different temporal resolutions,  be adapted for best accuracy performance under federated learning?}} 

\section{SSM-based neuron models}
\label{sec:SSM-based-NN} 

For our deep neural networks models, we study two neuron model families: \textbf{LIF} spiking neurons, chosen for their widespread use in spiking neural networks and  hence neuromorphic computing, and \textbf{SSMs with linear state transitions}, chosen as a non-spiking baseline for comparison. For each neuron model, we consider two variants:
\begin{itemize}
    \item A \textbf{\normal}variant that trains the dynamical parameters directly.
    \item A \textbf{$\boldsymbol{\Delta}$-}variant that adopts an exp-log reparameterization that maps the local temporal resolution $T_k$ to an explicitly trainable neuron temporal resolution parameter $\Delta$.
\end{itemize}
All neurons within a given network share the same model.

\subsection{Leaky Integrate-and-Fire (LIF) model}
\label{sec:lif_model}
The discrete-time Leaky Integrate-and-Fire (LIF) neuron in a feed-forward SNN is defined as in~\cite{bittar2022surrogate}:
\begin{subequations}
\label{eqn:LIFneuron}
\begin{align}
    \Umem{t+1} &= \alpha \bigl(\Umem{t} - \vartheta \Sil{t}\bigr) + (1-\alpha)\,u[t], \\
    \Sil{t}    &= h_\vartheta \bigl(\Umem{t}\bigr), \label{eqn:LIF_spk}
\end{align}
\end{subequations}
where all variables are scalar-valued, $u[t] \in \mathbb{R}$ is the input, $\Umem{t} \in \mathbb{R}$ is the membrane potential, $\Sil{t} \in \{0,1\}$ is the output spike produced by the Heaviside step function $h_\vartheta(\cdot)$ with threshold $\vartheta$, and $\alpha \in (0,1)$ controls the decay of the membrane potential. The below two variants differ only in how $\alpha$ is parametrised:

\textbf{\Normal-LIF:} The parameter $\alpha$ is trained directly as a free scalar parameter.

\textbf{$\Delta$-LIF:} 
In a standard LIF neuron, $\alpha$ represents the decay of the membrane potential and it is often parametrized as $\alpha=\exp(-\Delta \gamma)$  where $\Delta$ is associated with  the duration of a timestep, i.e. time-resolution,  and $\gamma$ is the reciprocal of the membrane time constant. 
Following~\cite{fabre2026silifstructuredstatespace}, we instead express $\alpha$ as 
$\alpha = \exp\!\bigl(-e^{\Delta_{\log}}\,e^{\gamma_{\log}}\bigr),$
where $\Delta_{\log}$ and $\gamma_{\log}$ are trained as separate parameters.

\subsection{Linear-Dynamics Diagonal State Space Model (LD-SSM)}
\label{sec:time-invariant-ssm}
A time-invariant  discrete-time SSM with linear state-transition dynamics  is given by~\cite{ljung_SI}:
\begin{subequations}
\label{eqn:SSM:TInv}
\begin{align}
    \xbf[t+1] &= \Abf\,\xbf[t] + \Bbf\,\ubf[t], \label{eqn:SSM:TInv:state}\\
    \ybf[t]   &= h(\Cbf\,\xbf[t]) \label{eqn:SSM:TInv:output}
\end{align} 
\end{subequations}
with state transition matrix $\Abf \in \mathbb{C}^{N \times N}$, input matrix $\Bbf \in \mathbb{C}^{N \times n_{\mathrm{in}}}$, and output matrix $\Cbf \in \mathbb{C}^{n_{\mathrm{out}} \times N}$. The neuron output is obtained by applying a continuous valued activation function $h(\cdot)$. 
Following \cite{gu2022efficientlymodelinglongsequences}, we consider the case where the state transition matrix is diagonal.  

\textbf{\Normal-SSM:} All three matrices $\Abf$, $\Bbf$, $\Cbf$ are trained directly.

\textbf{$\Delta$-SSM:} Inspired by the parametrisation in~\cite{gu2022efficientlymodelinglongsequences} code implementation, the diagonal entries of $\Abf$ are expressed as
$
    \Abf = \exp\!\bigl(-e^{\Delta_{\log}}\,(e^{\Abf_{\Re,\log}} - j\,\Abf_{\Im})\bigr),
$
where $\Abf_{\Re,\log}, \Abf_{\Im} \in \mathbb{R}^{N \times N}$ are diagonal matrices and $\Delta_{\log} \in \mathbb{R}$ is a scalar. The output matrix $\Cbf $ is obtained using a trainable $\tilde{\Cbf}$ which is rescaled as $\Cbf = \tilde{\Cbf} \cdot (e^{\Abf} - \Ibf)\Abf_c^{-1
}$, where $\Abf_c = e^{\Abf_{\Re,\log}} - j\,\Abf_{\Im}$. The parameters $\Abf_{\Re,\log}$, $\Abf_{\Im}$, and $\Delta_{\log}$ are trained instead of the entries of $\Abf$, while $\Bbf$ and $\tilde{\Cbf}$ remain directly trained.

\paragraph{Remark.}
Linear state-transition dynamics combined with a continuous-valued activation (e.g. GELU) is not a SNN neuron model and it is not strictly neuromorphic-native, as it lacks the event-driven binary-spike communication that partially motivates the operating principles of neuromorphic hardware. We nonetheless include \normal-SSM and $\Delta$-SSM as baselines for their spiking counterparts, given the recent strong performance of deep SSM architectures~\cite{gu2022efficientlymodelinglongsequences}.
% https://github.com/state-spaces/s4/blob/main/models/s4/s4d.py look at lines 14-46

\section{Proposed Framework}
\label{sec:proposed_framework}

\subsection{Proposed Algorithm Structure}\label{sec:alg}

We adopt \textsc{FedAvg}~\cite{mcmahan2017communication} as the backbone of our federated learning procedure, and integrate an adaptation mechanism into the loop to address heterogeneous temporal resolutions. At each communication round, before aggregation, each local model $\thetabf_k = \{\phibf_{k},\, \Wbf_k,\, \psibf_k\}$ is adapted from its client-specific temporal resolution $T_k$ to a user-defined server resolution $\Tscentral$; symmetrically, before the updated global model is dispatched to client $k$, it is adapted back from $\Tscentral$ to $T_k$. The adaptation is applied \textbf{only} to $\phibf$: the synaptic weights $\Wbf$ and the normalization statistics $\psibf$ are not altered  during this adaptation. 
The specific adaptation rule depends on the neuron model and is detailed in Section~\ref{sec:adaptations}. The overall  Federated Learning with Time Resolution Adaptation (FedTA) framework is summarized in Algorithm~\ref{alg:main_framework}. See Figure \ref{fig:FL_set_up:detailed} for visualization of FedTA.

\begin{algorithm}[t]
\caption{FedTA: Federated learning with temporal resolution adaptation.}
\label{alg:main_framework}
\DontPrintSemicolon
\SetKwInOut{Input}{Input}\SetKwInOut{Output}{Output}
\BlankLine
\textbf{(1) Initialize global model}\;
Randomly initialize
$\thetabf_c = \{\phibf_{c},\, \Wbf_c,\, \psibf_c\}$
\BlankLine
\For{round $= 1, \dots, R$}{
\textbf{(2) Centrally adapt to clients and broadcast}\;
\For{each client $k = 1, \dots, K$}{
    Adapt neuron dynamics at central server:
    $\phibf_{c}
     \xrightarrow{\Tscentral \to T_k}
     \phibf_{k}$\;
    Send $\thetabf_k =
    \{\phibf_{k},\, \Wbf_c,\, \psibf_c\}$
    to client $k$\;
}
\BlankLine
\textbf{(3) Train locally}\;
\For{each client $k = 1, \dots, K$}{
    \For{epoch $= 1, \dots, E$}{
        Update $\thetabf_k$ via optimizer on $\Dcal_k$\;
    }
}
\Return Updated local models $\{\thetabf_k\}_{k=1}^K$ to central server\;
\BlankLine
\textbf{(4) Centrally adapt and aggregate}\;
\For{each client $k = 1, \dots, K$}{
    Adapt neuron dynamics at central server:
    $\phibf_{k}
     \xrightarrow{T_k \to \Tscentral}
     \phibf_{c,k}$\;
    $\thetabf_k = \{\phibf_{c,k},\, \Wbf_k,\, \psibf_k\}$\;
}
Aggregate via \textsc{FedAvg}:
$\thetabf_c \leftarrow \frac{1}{K}\sum_{k=1}^{K} \thetabf_k$\;
}
\BlankLine
\textbf{(5) Evaluate}\;
\Input{Global model $\thetabf_c$, held-out test set $\Dcal_{\mathrm{test}}$}
\Output{Performance of $\thetabf_c$ on $\Dcal_{\mathrm{test}}$}
\end{algorithm}

\subsection{Dynamics Adaptation Methods}
\label{sec:adaptations}
We now present the dynamics adaptation methods that we propose to integrate in FedTA~\ref{alg:main_framework}. 
In Section~\ref{sec:neuronModelAdaptation} we detail how each adaptation is concretely applied to the dynamics parameters $\phibf$ of each neuron model in Section \ref{sec:SSM-based-NN}.

\textbf{Integral and Euler Temporal Resolution Adaptation Methods:}
Building on \cite{karilanova2025zeroshottemporalresolutiondomain} we consider two adaptation methods derived for linear SSMs that operate on $(\Abf, \Bbf)$. Given a discrete-time SSM  whose parameters are denoted with subscript $r_1$ and associated with the temporal resolution $\ctR{1}$, we would like to adapt the parameters of the SSM for the temporal resolution $\ctR{2}$. We  denote the resulting SSM  associated with $\ctR{2}$ with the subscript $r_2$. Let $\DeltaRatio=\frac{\ctR{2}}{\ctR{1}}$.

The {\it{Euler temporal adaptation method}} is given by: 
\begin{subequations}
\label{eqn:integral_adaptation_method}
\begin{align} 
 \AbfR{2} &= \Ibf + \DeltaRatio (\AbfR{1} - \Ibf) \\
 \BbfR{2} &= \DeltaRatio  \BbfR{1} 
\end{align}
\end{subequations}
Incorporating the above into Algorithm~\ref{alg:main_framework}, we obtain the proposed algorithm FedTA-Eul.  

The {\it{Integral temporal adaptation method}} is given by:
\begin{subequations}
\label{eqn:euler_adaptation_method}
\begin{align} 
 \! \AbfR{2} &= \AbfR{1} ^ \DeltaRatio \\
 \! \BbfR{2} &= (\AbfR{2} - \Ibf) (\AbfR{1} - \Ibf)^{-1} \BbfR{1} 
 \end{align}
\end{subequations}
Incorporating the above into Algorithm~\ref{alg:main_framework}, we obtain the proposed algorithm FedTA-Int.

\textbf{$\Delta$-Temporal Resolution Adaptation Method}
For model parameters in which the temporal resolution is captured through an explicitly trained parameter $\Delta_{\log} \equiv \ln(\Delta)$, adapting from $\Delta_{r_1}$ to $\Delta_{r_2} = \rho\,\Delta_{r_1}$ reduces to an additive shift:
\begin{align}
    \Delta_{\log,r_2} = \Delta_{\log,r_1} + \ln(\rho),
    \label{eqn:Delta_scaled_adaptation_method}
\end{align}
as $\ln(\Delta_{r_2}) = \ln(\rho\,\Delta_{r_1}) = \ln(\rho) + \ln(\Delta_{r_1})$.
No other parameters are modified. Combining this with Algorithm~\ref{alg:main_framework}, we obtain the proposed algorithm FedTA-$\Delta$.

\subsection{Neuron-Models under Adaptation Methods}\label{sec:neuronModelAdaptation}

\Normal-SSM:
The dynamics parameters are $\phi = \{\Abf, \Bbf, \Cbf\}$. The Integral and Euler adaptation methods formulas (\eqref{eqn:integral_adaptation_method} and \eqref{eqn:euler_adaptation_method} respectively) apply directly to $(\Abf, \Bbf)$. The output matrix $\Cbf$ is left unchanged.

$\Delta$-SSM:
The dynamics parameters are $\phi = \{\Abf_{\Re,\log},\, \Abf_{\Im},\, \Delta_{\log},\, \Bbf,\, \tilde{\Cbf}\}$. Since $\Delta_{\log}$ explicitly encodes the temporal resolution, the $\Delta$-method (\eqref{eqn:Delta_scaled_adaptation_method}) is applied to $\Delta_{\log}$ while all other parameters are left unchanged.

\Normal-LIF:
The dynamics parameter is $\phi = \{\alpha\}$.
Comparing~\eqref{eqn:LIFneuron} with the linear SSM~\eqref{eqn:SSM:TInv}, $\alpha$ plays the role of the (scalar, real) state-transition parameter $\Abf$.
We therefore apply the Integral and Euler formulas to $\alpha$, treating it as a $1{\times}1$ real version of $\Abf$.
This analogy is imperfect: the spike-reset term $-\alpha\theta S[t]$ in~\eqref{eqn:LIFneuron} introduces a nonlinearity absent from linear SSMs, so the adaptation is an approximation rather than an exact change of temporal resolution. However as illustrated in \cite{karilanova2025zeroshottemporalresolutiondomain} even though it is an approximation, it can be used to partially compensate for temporal mismatch.

$\Delta$-LIF:
The dynamics parameters are $\phi = \{\Delta_{\log},\, \gamma_{\log}\}$.
%where $\alpha = \exp(-e^{\Delta_{\log}} e^{\gamma_{\log}})$. 
%
As in $\Delta$-SSM, $\Delta_{\log}$ explicitly encodes the temporal resolution,
so the $\Delta$ method applies directly: $\Delta_{\log}$ is shifted by $\ln(\rho)$ while $\gamma_{\log}$ is left unchanged.

\section{Numerical Experiments Set up}\label{sec:num:setting:maintext}

\textbf{Data:} 
We use two popular neuromorphic datasets. The Spiking Heidelberg Digits (SHD) dataset \cite{shddataset} contains spoken digits $(0\!-\!9)$ in both English and German, yielding $20$ classes split across $8156$ training and $2264$ testing samples. The DVS-Gesture dataset \cite{Amir_2017_CVPR} contains event-based recordings of $11$ distinct hand gestures split into $1086$ training and $256$ testing samples. Both datasets are used for classification tasks. Additional information on the datasets and their processing is presented in Section~\ref{sec:datasets_details}. The training data was partitioned across clients in an independent and identically distributed (IID) manner with respect to class labels, ensuring that each client observes a balanced subset of all classes.

\textbf{Network Architecture and Training:}
% Code/training details
All models were optimized using AdamW \cite{loshchilov2019decoupledweightdecayregularization} together with a cosine learning rate schedule. For the normalization layers we used Batch Normalization \cite{ioffe2015batchnormalizationacceleratingdeep}. Additional information such as initilization of parameters and used hyperparameters is provided in Section~\ref{sec:details_network_and_training}.

\section{Results}
\label{sec:results}

% ============================================================
%  TABLE: Scenario A and Scenario B
% ============================================================
\begin{table}[t]
\centering
\caption{Test accuracy on clients with heterogeneous temporal resolution. Averages over $5$ runs. Shown in bold are best values per row. Scenario~A has one client per resolution and Scenario~B has five clients per resolution (3 and 15 clients total), across temporal resolutions $\Tsnode \in \{1, 2, 4\}$.
}
\label{tab:heterogeneous_Ts}
\newcolumntype{M}{>{\centering\arraybackslash}p{3.5cm}}
\newcolumntype{K}{>{\centering\arraybackslash}p{0.4cm}}

% --- SHD subtable ---
\begin{subtable}{\linewidth}
\caption{SHD dataset.}
\label{tab:all_shd}
\resizebox{\textwidth}{!}{%
\begin{tabular}{MK ccc cc ccc cc}
\toprule
\multirow{2}{*}{Clients configuration}
  & \multirow{2}{*}{$\Tscentral$}
  & \multicolumn{3}{c}{\normal-LIF}
  & \multicolumn{2}{c}{$\Delta$-LIF}
  & \multicolumn{3}{c}{\normal-SSM}
  & \multicolumn{2}{c}{$\Delta$-SSM} \\
\cmidrule(lr){3-5} \cmidrule(lr){6-7} \cmidrule(lr){8-10} \cmidrule(lr){11-12}
 & & FedAvg & FedTA-Int & FedTA-Eul
   & FedAvg & FedTA-$\Delta$
   & FedAvg & FedTA-Int & FedTA-Eul
   & FedAvg & FedTA-$\Delta$ \\
\midrule
\multirow{3}{*}{Scenario A} & 1 & 76.7{\tiny$\pm$0.9} & 75.6{\tiny$\pm$1.9} & 75.4{\tiny$\pm$1.4} & 80.2{\tiny$\pm$0.9} & 65.3{\tiny$\pm$2.9} & 83.8{\tiny$\pm$1.1} & \textbf{91.6{\tiny$\pm$0.6}} & 60.7{\tiny$\pm$2.0} & 79.8{\tiny$\pm$0.8} & 86.5{\tiny$\pm$1.8} \\
 & 2 & 79.2{\tiny$\pm$0.1} & 78.9{\tiny$\pm$0.6} & 78.9{\tiny$\pm$0.6} & 83.1{\tiny$\pm$0.6} & 83.3{\tiny$\pm$0.4} & 87.4{\tiny$\pm$0.9} & \textbf{90.2{\tiny$\pm$0.6}} & 87.6{\tiny$\pm$1.7} & 88.7{\tiny$\pm$0.9} & 87.8{\tiny$\pm$0.8} \\
 & 4 & 65.6{\tiny$\pm$0.5} & 61.2{\tiny$\pm$1.5} & 60.8{\tiny$\pm$1.7} & 72.6{\tiny$\pm$0.6} & 62.8{\tiny$\pm$1.8} & 73.7{\tiny$\pm$1.9} & \textbf{84.3{\tiny$\pm$1.1}} & 65.2{\tiny$\pm$1.6} & 74.3{\tiny$\pm$1.8} & 32.9{\tiny$\pm$5.8} \\
\cmidrule(lr){1-12}
\multirow{3}{*}{Scenario B} & 1 & 63.3{\tiny$\pm$1.9} & 58.4{\tiny$\pm$1.8} & 58.3{\tiny$\pm$1.6} & 70.9{\tiny$\pm$0.7} & 51.6{\tiny$\pm$1.2} & 82.4{\tiny$\pm$2.8} & \textbf{89.9{\tiny$\pm$0.9}} & 50.0{\tiny$\pm$3.9} & 73.4{\tiny$\pm$3.2} & 79.8{\tiny$\pm$2.5} \\
 & 2 & 67.5{\tiny$\pm$1.1} & 55.6{\tiny$\pm$0.9} & 54.7{\tiny$\pm$1.4} & 75.5{\tiny$\pm$0.7} & 74.5{\tiny$\pm$0.9} & 85.1{\tiny$\pm$0.5} & \textbf{88.5{\tiny$\pm$0.6}} & 80.9{\tiny$\pm$1.1} & 78.3{\tiny$\pm$1.8} & 78.0{\tiny$\pm$2.2} \\
 & 4 & 51.1{\tiny$\pm$1.8} & 41.0{\tiny$\pm$2.2} & 40.4{\tiny$\pm$1.6} & 62.6{\tiny$\pm$1.1} & 45.7{\tiny$\pm$1.1} & 64.9{\tiny$\pm$2.5} & \textbf{81.7{\tiny$\pm$0.5}} & 59.0{\tiny$\pm$1.9} & 47.7{\tiny$\pm$2.9} & 26.8{\tiny$\pm$6.7} \\
\bottomrule
\end{tabular}}
\end{subtable}

\vspace{1em}

% --- DVSGesture subtable ---
\begin{subtable}{\linewidth}
\caption{DVSGesture dataset.}
\label{tab:all_dvs}
\resizebox{\textwidth}{!}{%
\begin{tabular}{MK ccc cc ccc cc}
\toprule
\multirow{2}{*}{Clients configuration}
  & \multirow{2}{*}{$\Tscentral$}
  & \multicolumn{3}{c}{\normal-LIF}
  & \multicolumn{2}{c}{$\Delta$-LIF}
  & \multicolumn{3}{c}{\normal-SSM}
  & \multicolumn{2}{c}{$\Delta$-SSM} \\
\cmidrule(lr){3-5} \cmidrule(lr){6-7} \cmidrule(lr){8-10} \cmidrule(lr){11-12}
 & & FedAvg & FedTA-Int & FedTA-Eul
   & FedAvg & FedTA-$\Delta$
   & FedAvg & FedTA-Int & FedTA-Eul
   & FedAvg & FedTA-$\Delta$ \\
\midrule
\multirow{3}{*}{Scenario A} & 1 & 83.4{\tiny$\pm$1.4} & 82.1{\tiny$\pm$1.8} & 82.4{\tiny$\pm$2.3} & 87.5{\tiny$\pm$1.9} & 86.6{\tiny$\pm$1.4} & 79.3{\tiny$\pm$6.9} & \textbf{91.1{\tiny$\pm$1.1}} & 88.0{\tiny$\pm$0.6} & 62.9{\tiny$\pm$14.1} & 75.2{\tiny$\pm$2.0} \\
 & 2 & 82.0{\tiny$\pm$1.2} & 78.1{\tiny$\pm$1.4} & 79.9{\tiny$\pm$1.8} & 87.7{\tiny$\pm$1.3} & 87.6{\tiny$\pm$0.6} & 87.8{\tiny$\pm$0.5} & \textbf{89.3{\tiny$\pm$1.7}} & 85.2{\tiny$\pm$1.9} & 70.4{\tiny$\pm$8.1} & 75.5{\tiny$\pm$1.8} \\
 & 4 & 77.3{\tiny$\pm$2.1} & 71.7{\tiny$\pm$2.5} & 71.1{\tiny$\pm$1.0} & \textbf{85.2{\tiny$\pm$0.6}} & 79.8{\tiny$\pm$1.2} & 77.3{\tiny$\pm$2.3} & 77.3{\tiny$\pm$2.4} & 75.4{\tiny$\pm$2.3} & 59.2{\tiny$\pm$12.1} & 43.0{\tiny$\pm$10.7} \\
\cmidrule(lr){1-12}
\multirow{3}{*}{Scenario B} & 1 & 80.3{\tiny$\pm$2.0} & 78.2{\tiny$\pm$1.5} & 76.3{\tiny$\pm$3.1} & 87.7{\tiny$\pm$1.1} & 85.4{\tiny$\pm$1.6} & 79.9{\tiny$\pm$6.0} & \textbf{88.8{\tiny$\pm$0.7}} & 79.7{\tiny$\pm$1.5} & 70.9{\tiny$\pm$3.9} & 75.0{\tiny$\pm$2.1} \\
 & 2 & 78.7{\tiny$\pm$1.8} & 74.5{\tiny$\pm$2.5} & 73.9{\tiny$\pm$1.4} & \textbf{87.4{\tiny$\pm$1.8}} & 85.8{\tiny$\pm$0.9} & 77.2{\tiny$\pm$11.4} & 82.6{\tiny$\pm$4.3} & 76.6{\tiny$\pm$1.8} & 71.8{\tiny$\pm$3.7} & 72.3{\tiny$\pm$4.9} \\
 & 4 & 74.7{\tiny$\pm$1.8} & 70.2{\tiny$\pm$1.9} & 69.3{\tiny$\pm$1.9} & \textbf{83.8{\tiny$\pm$1.3}} & 75.8{\tiny$\pm$1.9} & 63.4{\tiny$\pm$11.0} & 74.1{\tiny$\pm$2.6} & 73.9{\tiny$\pm$0.7} & 44.3{\tiny$\pm$19.1} & 22.7{\tiny$\pm$4.1} \\
\bottomrule
\end{tabular}}
\end{subtable}

\end{table}

\subsection{Heterogeneous temporal resolution across clients}
\label{sec:hetero_temporal_res_results}

We explore two different client scenarios, Scenario~A and~B, see Table~\ref{tab:heterogeneous_Ts}. In both scenarios, clients operate at one of three temporal resolutions $\Tsnode \in \{1, 2, 4\}$, where $\Tsnode=1$ corresponds to the finest temporal resolution, $\Tsnode=4$ to the coarsest,
and $\Tsnode=2$ lies in between. Scenario~A has one client per resolution (3 clients total), while Scenario~B has five clients per resolution (15 clients total).

From Table~\ref{tab:heterogeneous_Ts}, the \normal-SSM and FedTA-Int combination achieves the  highest accuracy across both scenarios and all $\Tscentral$ on the SHD dataset. For DVS-Gesture, this holds in half of the cases, while the remaining half obtain highest accuracy by the $\Delta$-LIF and FedAvg combination.

Now we compare the performance of the algorithms for each neuron model. 
% normal-LIF and normal-SSM
For the \normal-LIF neuron, all algorithms perform within one standard deviation of each other when $\Tscentral \in \{1, 2\}$, while at $\Tscentral = 4$, FedAvg outperforms the adaptation variants.
For \normal-SSM, the FedTA-Int method clearly outperforms both FedAvg and FedTA-Eul. 
This can be attributed to the nature of the two approximation methods: Euler is a first-order approximation and thus less accurate than the Integral method. Under the linear \normal-SSM dynamics, it is therefore expected that FedTA-Eul underperforms. For \normal-LIF, however, both methods are approximations of inherently nonlinear dynamics, making neither a good fit, this may contribute to their close performance. 
% \Delta-LIF and \Delta-SSM
For the $\Delta$-LIF neuron, the FedAvg either outperforms or is within one standard distribution of the FedTA-$\Delta$. For the $\Delta$-SSM on the other hand for $\Tscentral \in \{1, 2\}$ the FedTA-$\Delta$ either outperforms or is within one standard distribution of the FedAvg, while for $\Tscentral=4$ the FedAvg clearly outperforms FedTA-$\Delta$.

We now focus on the effect of the centralized temporal resolution $\Tscentral$. Performance is generally highest at $\Tscentral = 1$ or $\Tscentral = 2$: the former preserves the finest temporal resolution and thus retains the most information, while the latter benefits from being the intermediate resolution in $\{1, 2, 4\}$, minimizing the average mismatch between the central and client feature spaces. 

Moving from Scenario~A ($3$ clients) to Scenario~B ($15$ clients) generally reduces accuracy, consistent with the increased difficulty of aggregation across a larger number of heterogeneous clients.

\subsection{Computational complexity of algorithms}
\label{sec:computational_complexity}

\begin{figure}
    \centering
    \caption{Computational complexity of the proposed algorithms on the DVSGesture dataset}
    \begin{subfigure}[b]{1\linewidth}
        \centering
        \caption{SHD dataset}
        \includegraphics[width=1\linewidth]{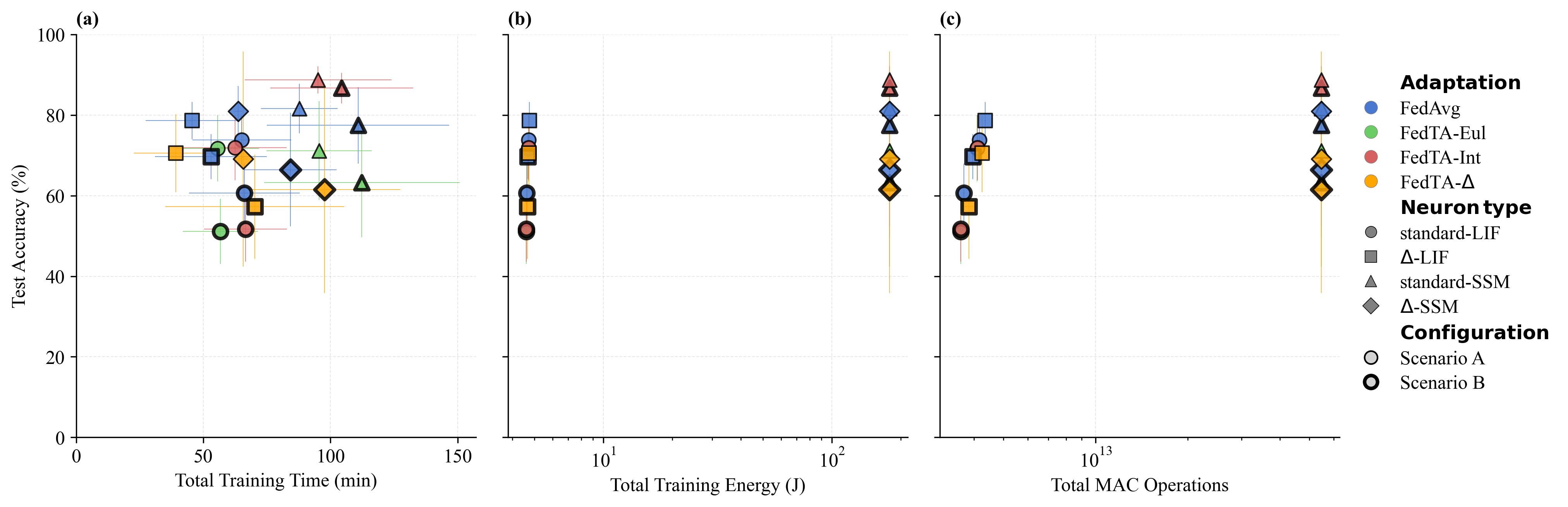}
        \label{fig:comp_complex_SHD}
    \end{subfigure}
    \begin{subfigure}[b]{1\linewidth}
        \centering
        \caption{DVSGesture dataset}
        \includegraphics[width=1\linewidth]{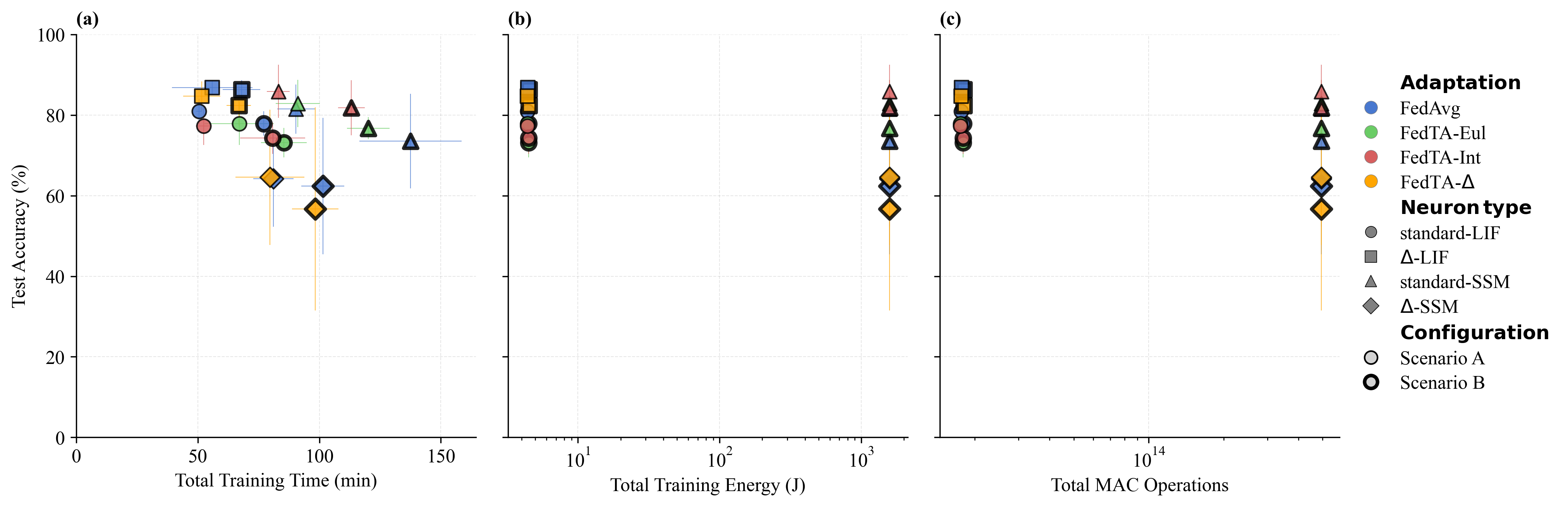}
        \label{fig:comp_complex_DVS}
    \end{subfigure}
\end{figure}

In this section we evaluate the computational complexity of the proposed algorithms using three metrics: total training time, estimated training energy, and total MAC operations, as reported in Figures~\ref{fig:comp_complex_SHD} and~\ref{fig:comp_complex_DVS}. All models are trained on an NVIDIA Tesla T4 GPU with 16GB RAM. For further details on the calculations, see Section~\ref{sec:energy_calculations}.

The computational overhead of FedTA-Int, FedTA-Eul and FedTA-$\Delta$ is negligible compared to FedAvg. For example, the training times remain within one standard deviation of the FedAvg baseline for each neuron model.
Figure \ref{fig:comp_complex_SHD} illustrates that the high accuracy of the \normal-SSM neuron model with FedTA-Int (red triangle) comes at a significantly higher energy and MAC cost compared to the other methods.
The $\Delta$-LIF neuron with FedAvg offers a promising trade-off, achieving a $44\times$ reduction in energy and $2.1\times$ faster training time at the cost of only $10\%$ accuracy. This is consistent with the known energy efficiency of SNN-based neuron models such as LIF, and their sparse and binary activations which result in AC rather than MAC operations.
Figure \ref{fig:comp_complex_DVS} illustrates that FedAvg combined with $\Delta$-LIF achieves the highest accuracy while remaining among the most computationally efficient models in terms of both training time and energy.

\subsection{Homogeneous temporal resolution across clients with central mismatch}

% ============================================================
%  Figures: LIF, SiLIF, SSM, SiSSM — SHD and DVSGesture
% ============================================================
\begin{figure}
    \centering
    \caption{Test accuracy under homogeneous client temporal resolution $\Tsnode$.}
    \label{fig:homogenious_Ts}
    \begin{subfigure}{\linewidth}
        \includegraphics[width=\linewidth]{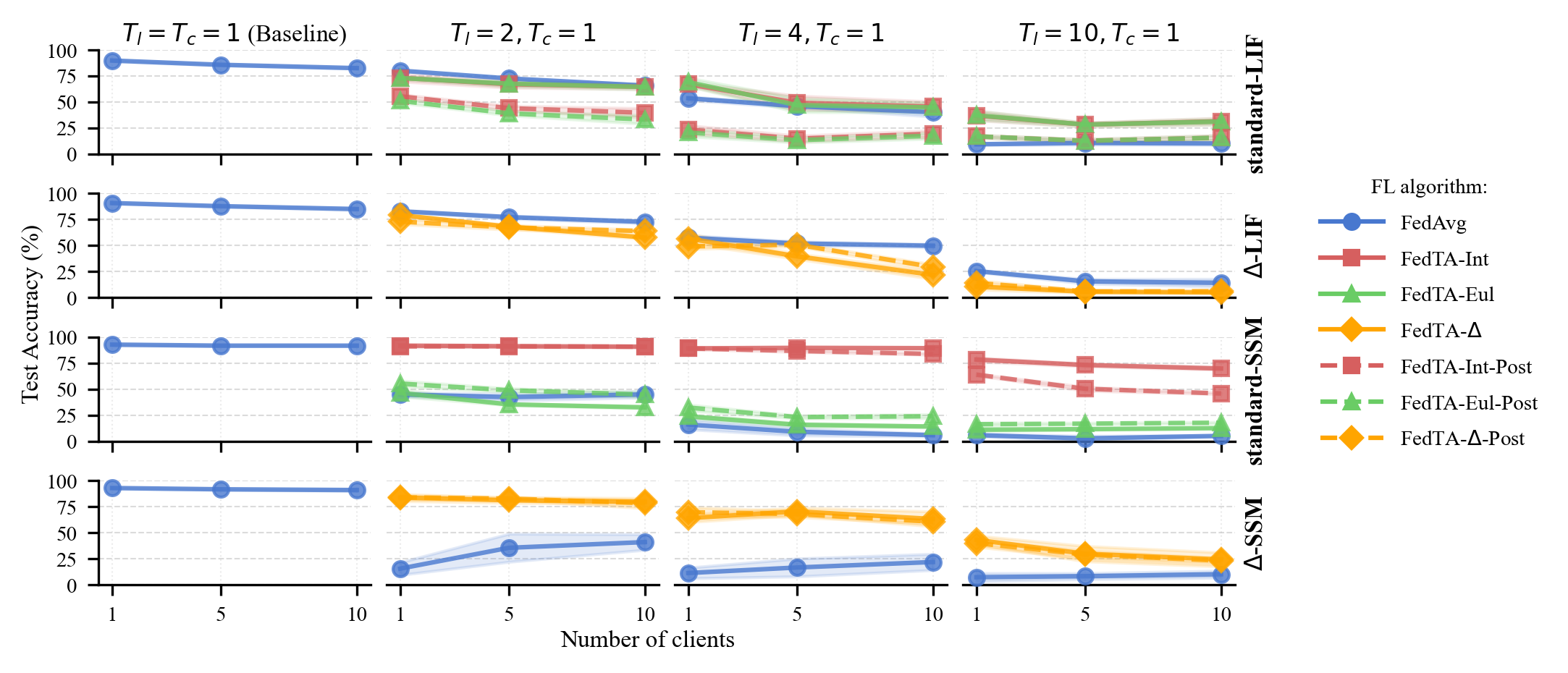}
        \caption{SHD Dataset}
    \end{subfigure}
    \begin{subfigure}{\linewidth}
        \includegraphics[width=\linewidth]{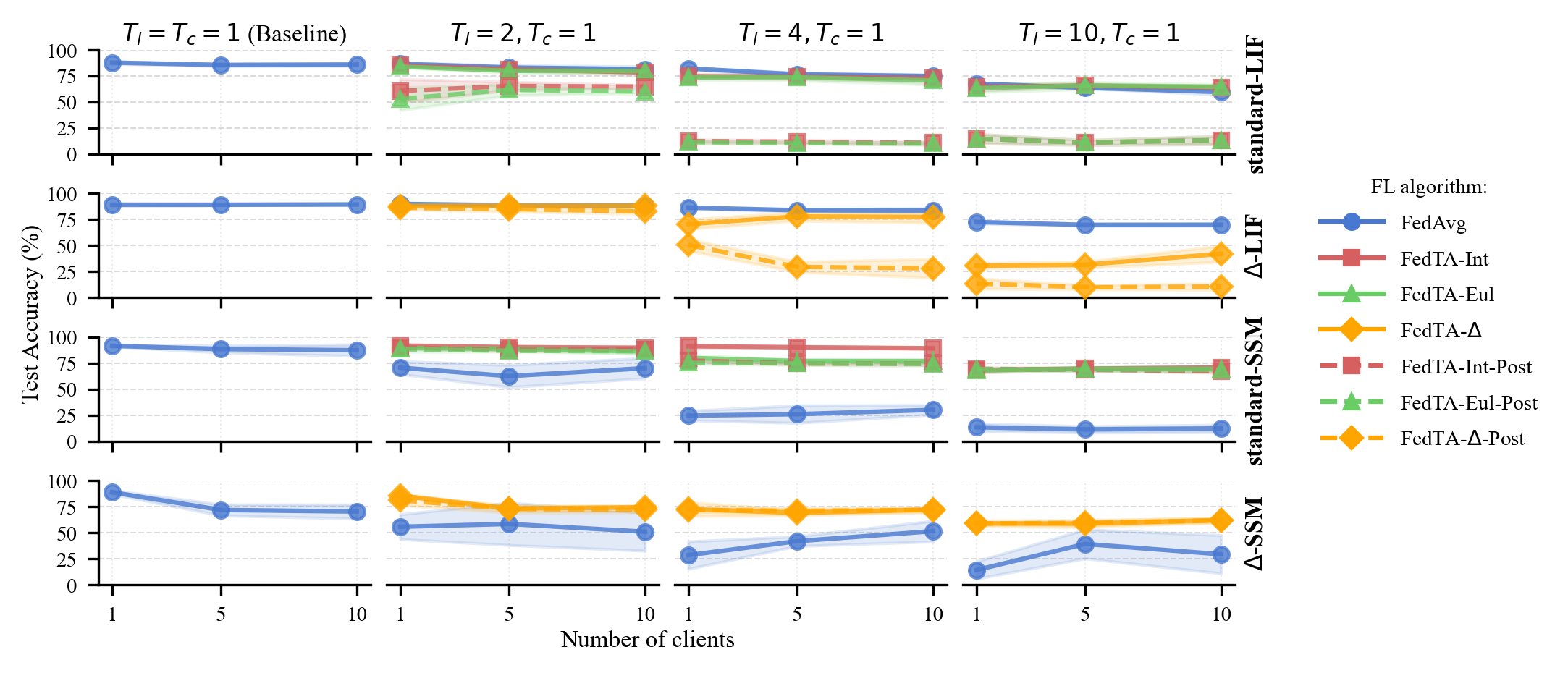}
        \caption{DVSGesture Dataset}
    \end{subfigure}
\end{figure}

We now consider the setting where all clients share the same temporal resolution $T_l$, which differs from the global model's resolution $T_c = 1$. This setting isolates to what extent performance loss arises from the center-to-client mismatch itself, disentangled from inter-client heterogeneity studied in Section~\ref{sec:hetero_temporal_res_results}. 
Since all clients are homogeneous here, adaptation can also be applied once post training at the final communication round rather than at each communication round. Therefore, we include three additional algorithms: FedTA-Int-Post, FedTA-Eul-Post, and FedTA-$\Delta$-Post.

We now analyse the robustness to temporal resolution mismatch of methods for each neuron model.
% General conclusion
At baseline ($T_l = T_c = 1$), all neuron types achieve above $89\%$ on SHD and $87\%$ on DVSGesture. As $T_l$ increases, accuracy degrades substantially for most neuron model and algorithm combinations. However, the degree of degradation and recovery varies by neuron type and FL algorithm used. 
% \normal-SSM
In particular, \normal-SSM suffers the sharpest decline under vanilla FedAvg but recovers strongly when paired with FedTA-Int, maintaining accuracy close to the $T_l = 1$ baseline. This result may explain the high performance of the \normal-SSM and FedTA-Int combination in the heterogeneous setting of Table~\ref{tab:heterogeneous_Ts}, i.e. being able to learn even from the clients operating at $T_l=4$. 
% \Delta-SSM
$\Delta$-SSM follows a similar pattern, recovering best under FedTA-$\Delta$. 
% LIF neurons
LIF-based models (\normal-LIF and $\Delta$-LIF) degrade more gradually under resolution mismatch but also benefit less from the proposed adaptation methods, suggesting their internal dynamics are inherently less sensitive to $T_l$ yet harder to correct when mismatch does occur.

For all proposed algorithms FedTA-Int, FedTA-Eul, FedTA-$\Delta$ it is either significantly better or almost equal to do adaptation at each step vs post training. This may be due to the fact that adapting at each round allows the global model to remain temporally consistent throughout training, rather than only correcting for the mismatch after convergence.

Increasing the number of clients from $1 \rightarrow 5 \rightarrow 10$ produces no consistent trend: accuracy occasionally degrades due to reduced per-client data, or improves slightly from implicit regularization, but remains largely insensitive to client count.

\section{Conclusions}
\label{sec:Discussion_and_Conclusions}

We proposed FedTA, a federated learning framework that addresses temporal resolution heterogeneity when clients collaboratively train neural networks with stateful neurons. FedTA integrates temporal adaptation into the federated averaging loop, allowing each client to train at its local resolution while maintaining a global stable model. We 
have integrated three dynamics adaptation methods into FedTA  yielding to FedTA-Int, FedTA-Eul, FedTA-$\Delta$, which we have evaluated across two neuromorphic benchmark datasets, two neuron model families and a range of time-resolution heterogeneity scenarios.

The FedTA-Int used with neurons with linear state dynamics and continuous-valued activation function achieves the highest accuracy in most scenarios, which is consistent with the match between the assumptions used in the adaptation method and these neuron models. On the other hand, 
LIF-based models, hence SNNs, are more robust across settings and offer a compelling energy-efficiency-accuracy trade-off. 
In all cases, the computation overhead of FedTA methods compared to vanilla model averaging is negligible. 
Together, these results highlight that FedTA is a promising framework for robust federated learning under temporal heterogeneity.

%  limitations 
FedTA-Int and FedTA-Eul rely on time-resolution adaptation methods  derived for linear SSM dynamics, making them limited approximations when applied to nonlinear dynamics, e.g. LIF neurons. Developing adaptation methods tailored to nonlinear neuron dynamics therefore represents an important avenue for future research.  Additionally, our framework focuses on heterogeneity in terms of temporal resolution, leaving other  forms of data heterogeneity, such as  class imbalance across clients, and their interactions  with temporal heterogeneity unexplored. Investigating these directions would further broaden the applicability of the framework.

\section*{Acknowledgment}
S. Karilanova acknowledges the support of Center for Interdisciplinary Mathematics (CIM), Uppsala University. 
The computations were enabled by resources provided by the National Academic Infrastructure for Supercomputing in Sweden (NAISS), partially funded by the Swedish Research Council through grant agreement no. 2022-06725.

\bibliographystyle{unsrt}
\bibliography{references}

\newpage
\section{Appendix}

\subsection{More Information on Network Architecture and Training}
\label{sec:details_network_and_training}
The state transition matrix $\Abf$ in the \normal-SSM and $\Delta$-SSM neuron models follows the diagonal structure of S4D~\cite{gu2022parameterization}, we use S4D-Lin scheme from the same work for parameter initialization.
We apply $h(\cdot) = \text{GELU}(\cdot)$ to the combined real and imaginary parts, i.e., $h(x) = \text{GELU}(\Re(x) + \Im(x))$.
The parameter $\alpha$ in the \normal-LIF is initialized uniformly at random in the interval
$\bigl[\exp(-\tfrac{1}{5}),\, \exp(-\tfrac{1}{25})\bigr]$, following~\cite{bittar2022surrogate}. For the $\Delta$-LIF neuron model we follow \cite{fabre2026silifstructuredstatespace} initialization. For LIF based neurons we use $\vartheta=1$ threshold.
All hyperparameters are reported in Table~\ref{tab:hyperparams}.

\begin{table}[]
    \centering
    \caption{Hyperparameters used for each dataset.}
    \label{tab:hyperparams}
    \newcolumntype{l}{>{\centering\arraybackslash}p{4cm}}
    \newcolumntype{k}{>{\centering\arraybackslash}p{4cm}}
    %----------------------------------------------------
    \begin{tabular}{lkk}
    \hline % 32768=(128*128*2)
    Parameter 
        & SHD
        & DVS-Gesture\\
    \hline
    $\cin$ & 140 & 32768 \\
    $\cout$ & 100 & 11 \\
    Nb. Hidden Layers & 2 & 3 \\
    $\Hneurons$ & 512 & 512 \\
    $\Nstate$ of SSM & 4 & 4 \\
    LR &$10^{-2}$& $10^{-2}$\\
    LR-SSM &$10^{-3}$ & $10^{-3}$\\
    WD  &$10^{-3}$ &$0.0$\\
    WD-SSM  &$10^{-3}$& $10^{-5}$\\
    Nb. Communication Rounds($=R$) & 20 & 100\\
    Nb. Local Epochs ($=E$) & 10 & 2\\
    Batch Size  & 64 & 16 \\
    Dropout  & 0.6 & 0.4 \\
    Window size ToFrame  & 10k & 80k \\
    \hline
    \end{tabular}
\end{table}

\subsection{More information on the data}
\label{sec:datasets_details}

% SHD
The Spiking Heidelberg Digits (SHD) dataset \cite{shddataset} is a spike-based audio benchmark produced by passing recordings through Lauscher, an artificial cochlea model that converts sound into unsigned spike trains. It contains spoken digits $(0\!-\!9)$ in both English and German, yielding $20$ classes split across $8156$ training and $2264$ testing samples, with each sample encoded over $700$ input channels. The current state-of-the-art on SHD reaches $95.1\%$ test accuracy, achieved by a spiking neural network with learnable synaptic delays \cite{hammouamri2023learningdelaysspikingneural}.

% DVS-Gesture
The DVS-Gesture dataset \cite{Amir_2017_CVPR} is among the most widely used benchmarks for event-based vision. It contains recordings of $29$ subjects performing $11$ distinct hand gestures under $3$ different lighting conditions, for a total of $1342$ samples ($1086$ for training and $256$ for testing). Recordings were captured with a $128 \times 128$ dynamic vision sensor, with two channels representing the positive and negative event polarities, and the task is to classify the gesture being performed. The strongest reported results to date are $99.3\%$ classification accuracy with a transformer-based model \cite{yao2023spike} and $97.7\%$ with a state-space model architecture \cite{Schone_EventbyEvent}.

For both datasets, spikes were binned into frames via the ToFrame transform from Tonic \cite{tonic}, using the time-window values listed in Table~\ref{tab:hyperparams}. These framed representations were then used to construct mini-batches during both training and evaluation.

To improve generalization, we applied a suite of data augmentations during training, namely \verb|TimeNeurons_mask_aug|, \verb|CutMix|, \verb|time_jitter_dense|, \verb|channel_jitter1d|, \verb|drop_by_time_dense|, and \verb|drop_by_channel_bloc|, together with additive noise. This augmentation strategy follows prior work \cite{deckers2024co, hammouamri2023learningdelaysspikingneural, Schone_EventbyEvent}.

For SHD specifically, we additionally applied spatio-temporal binning to reduce input dimensionality, consistent with previous studies \cite{hammouamri2023learningdelaysspikingneural, deckers2024co}. Groups of $5$ adjacent channels were merged, compressing the input from $700$ down to $140$ channels.

\subsection{Reporting errors}
\label{sec:report_errors}
All results are reported as mean $\pm$ standard deviation over $5$ independent runs. Each run produces a single accuracy value, computed as the mean accuracy over all test samples. The reported mean and standard deviation are taken across these $5$ values. In all tables and figures, we report one standard deviation.

\subsection{Energy Calculations}
\label{sec:energy_calculations}

We estimate energy consumption by calculating the number of multiply-accumulate (MAC) operations and their cost on conventional hardware, based on a 45\,nm CMOS process as described in~\cite{Horowitz}. This provides a hardware-independent basis for comparing computational cost across architectures i.e. it reflects the general cost of addition and multiplication in silicon, regardless of whether that silicon is organized as a GPU, CPU, or neuromorphic chip. This abstraction intentionally omits memory access patterns, data movement, and control logic, and is consistent with how energy efficiency is reported in the SNN literature~\cite{venkatesha2021federated, zhou2026sedformereventsynchronousspikingtransformers, 10.1007/978-3-031-30105-6_48}. Concretely, we adopt the costs of 3.1\,pJ per multiplication and 0.1\,pJ per addition.

We compute MAC operations for one forward pass through the network. Since the backward pass (backpropagation) is approximately twice as computationally expensive as the forward pass~\cite{wiedemann2020dithered, narayanan2021efficient}, the total cost per training step is approximately $3\times$ the forward-pass cost, accounting for one forward pass and two backward passes.

The reported MAC counts, and the energy estimates derived from them, cover the full training procedure: all time steps in each sequence, all training samples, all communication rounds, and all local epochs per round. Since each client operates at its own temporal resolution, the effective number of time steps per sample varies across clients; this is accounted for individually in our calculations. Operations involving complex-valued quantities are decomposed into their real-valued equivalents before counting.

For the LIF-based neuron models (\normal LIF and $\Delta$-LIF), the binary nature of spike activations means that the synaptic weight computation $\mathbf{W}\mathbf{s}$, where $\mathbf{s} \in \{0,1\}^n$, reduces to a sum of selected weight rows rather than a full matrix-vector multiplication. Accordingly, we count only additions for this operation, with no multiplications, consistent with standard practice in the SNN energy estimation literature~\cite{venkatesha2021federated}. Furthermore, since LIF neurons are frequently silent, we account for sparsity explicitly: for each client and each hidden layer, we measure the empirical spike rate and scale the addition count accordingly, so that only the fraction of time steps with active spikes contributes to the operation count.

\subsection{Accuracy vs Training Epochs}

\includegraphics[width=1\linewidth]{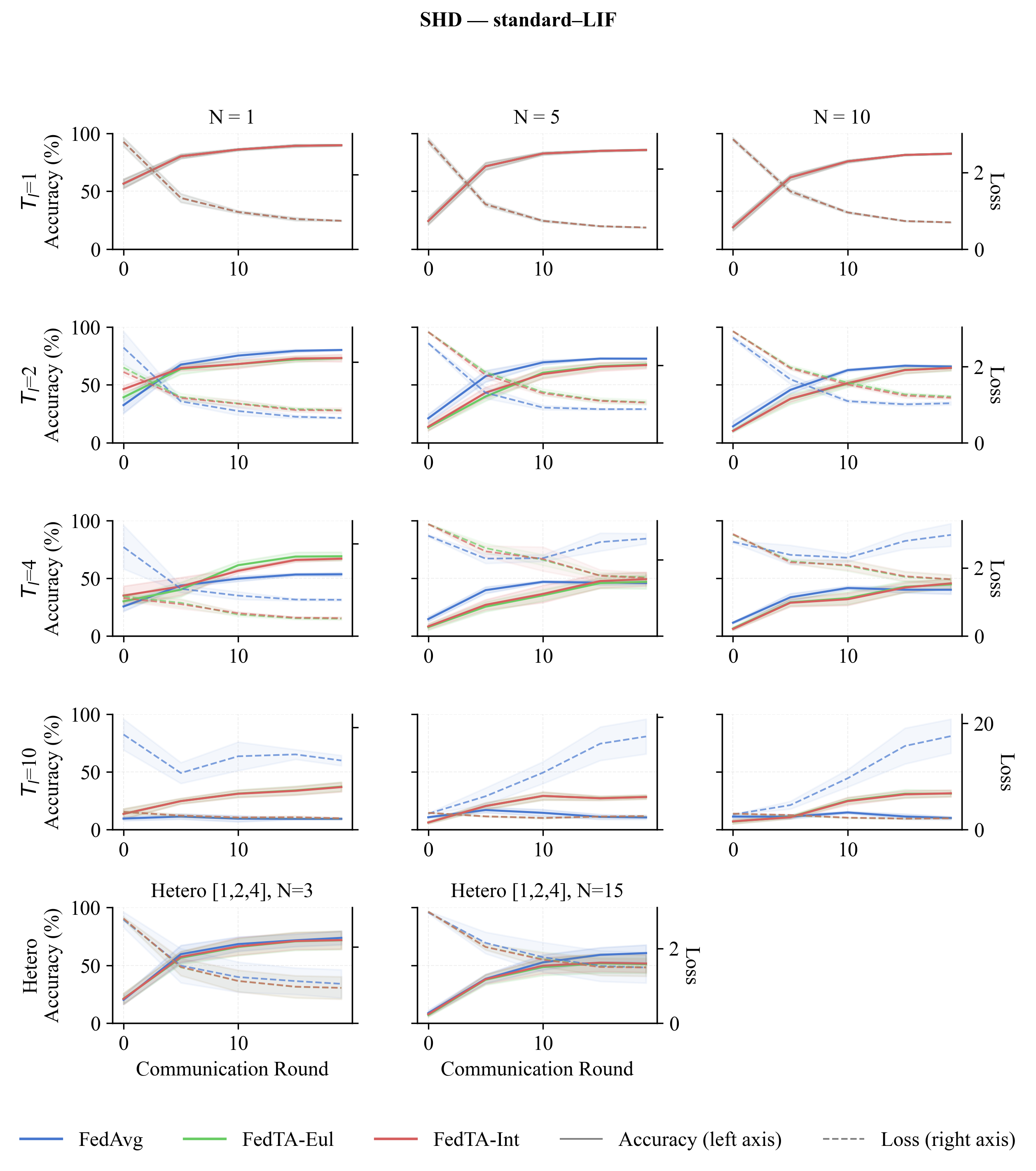}
\newpage
\includegraphics[width=1\linewidth]{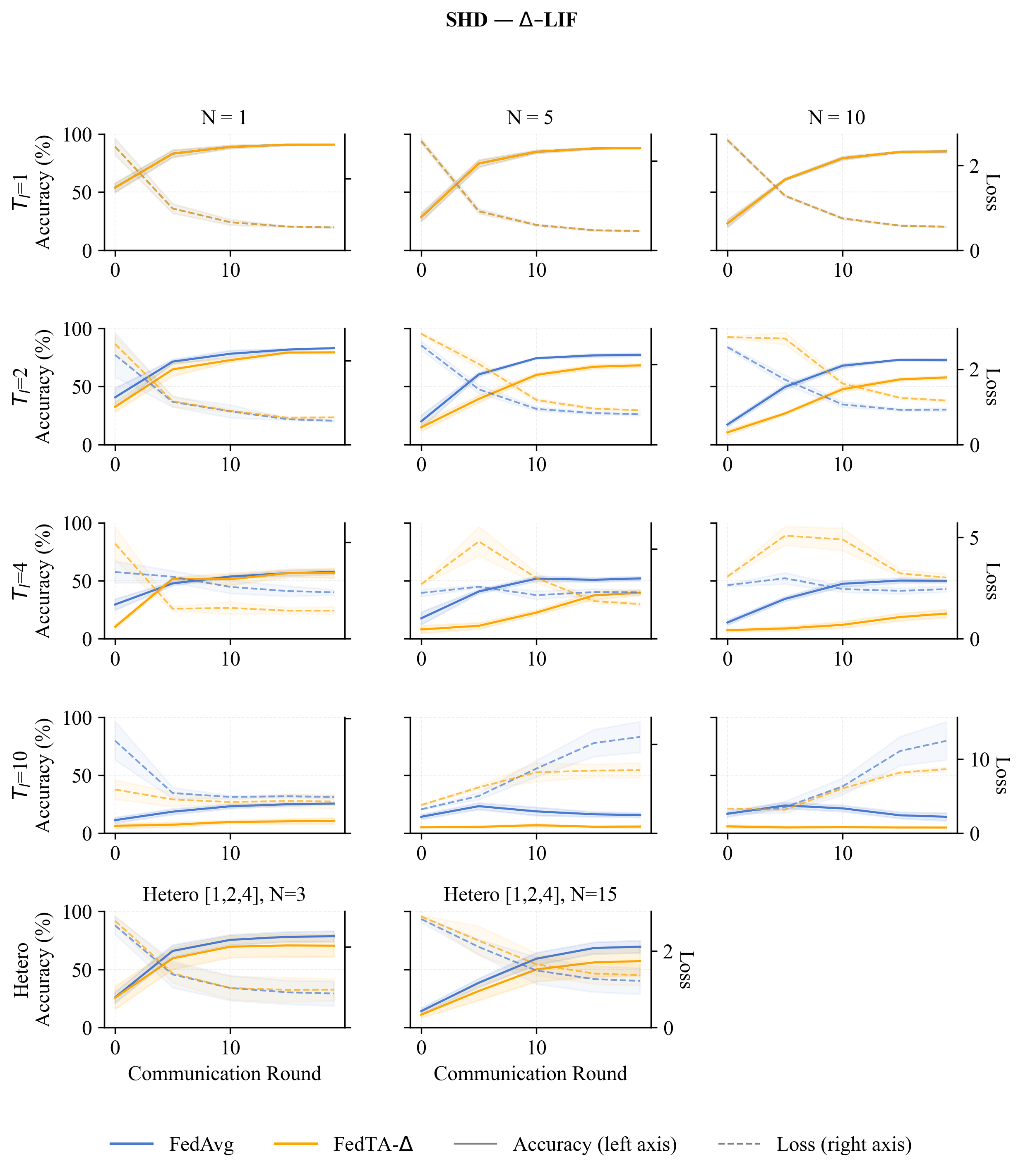}
\newpage
\includegraphics[width=1\linewidth]{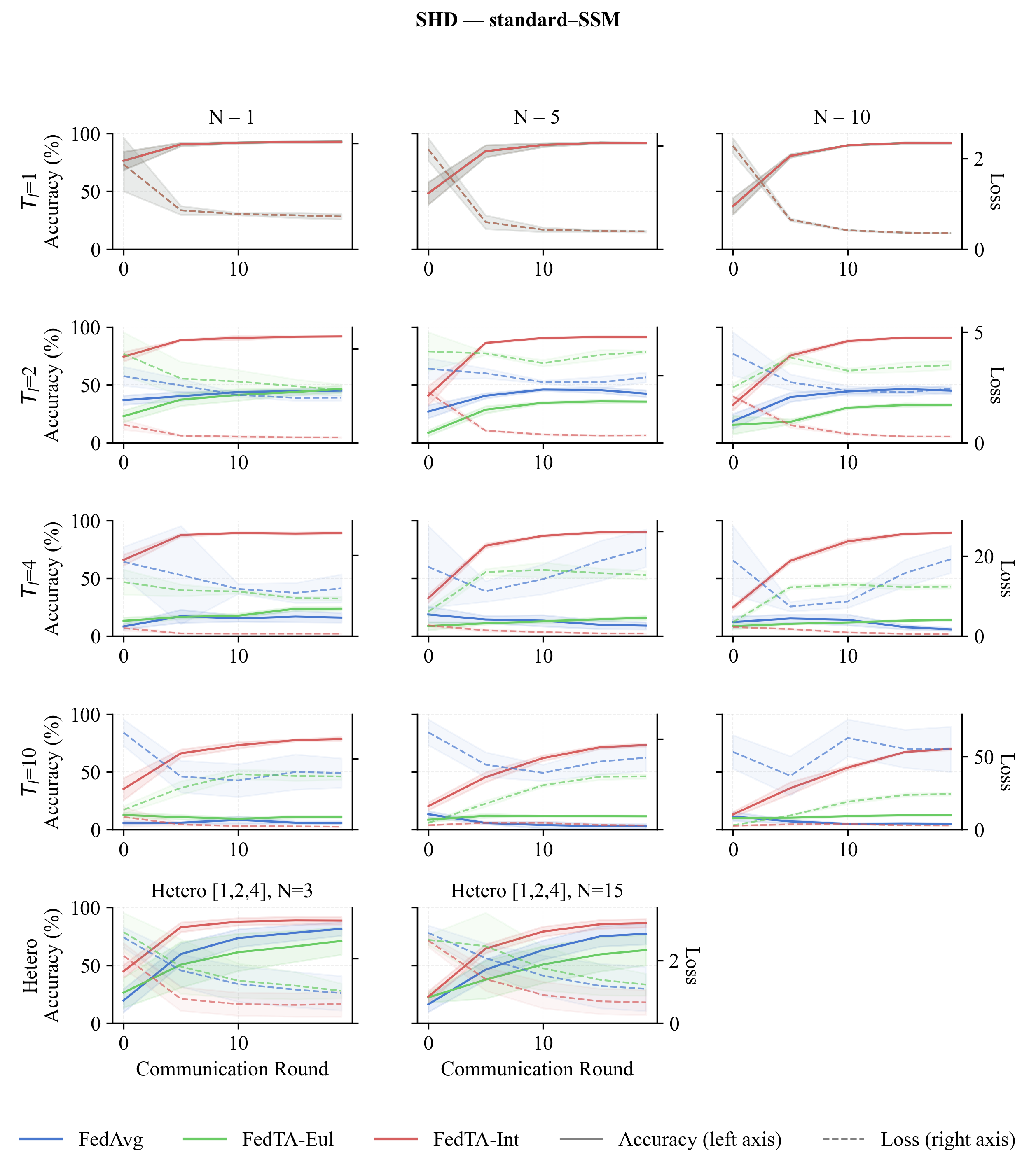}
\newpage
\includegraphics[width=1\linewidth]{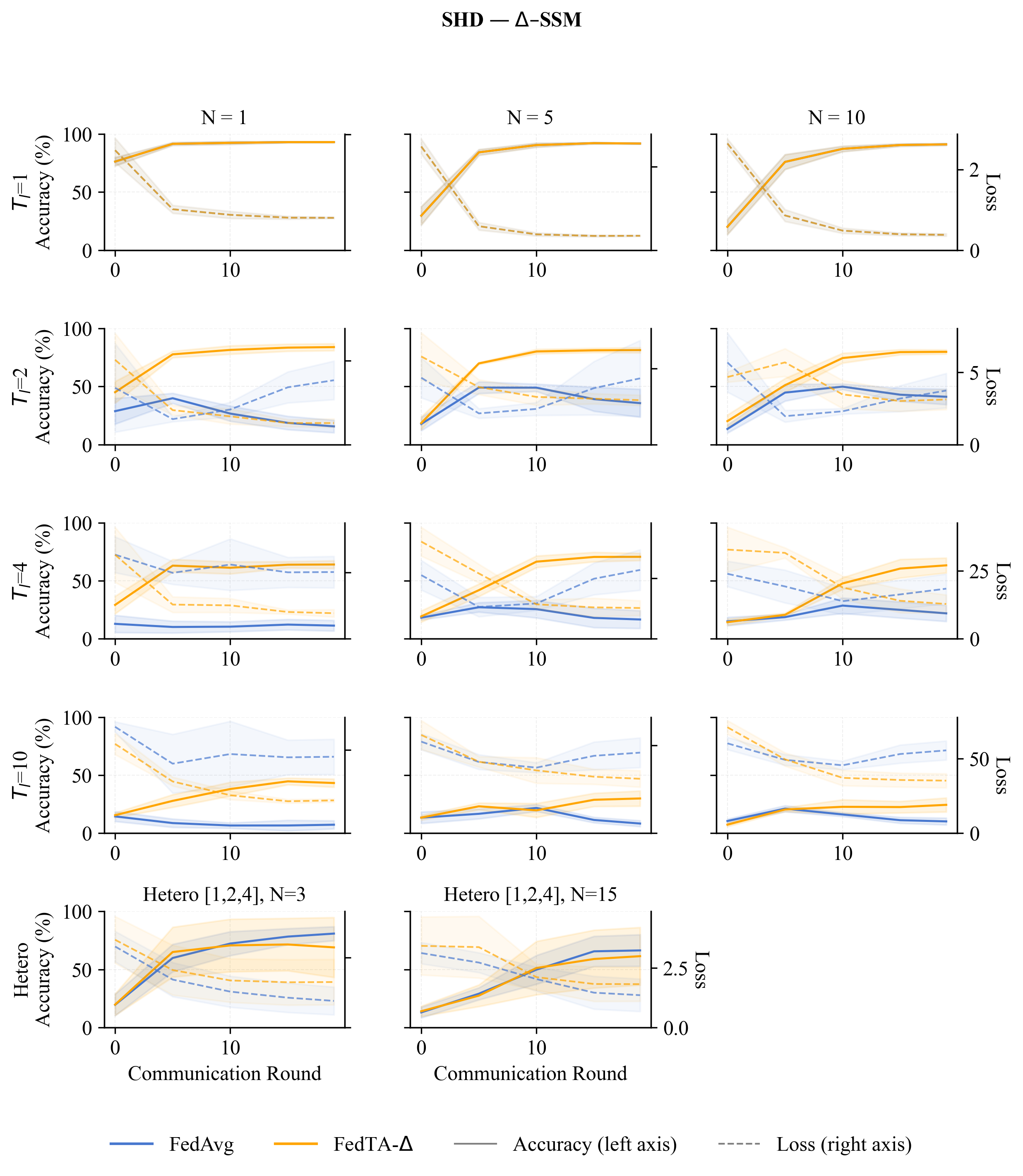}
\newpage
\includegraphics[width=1\linewidth]{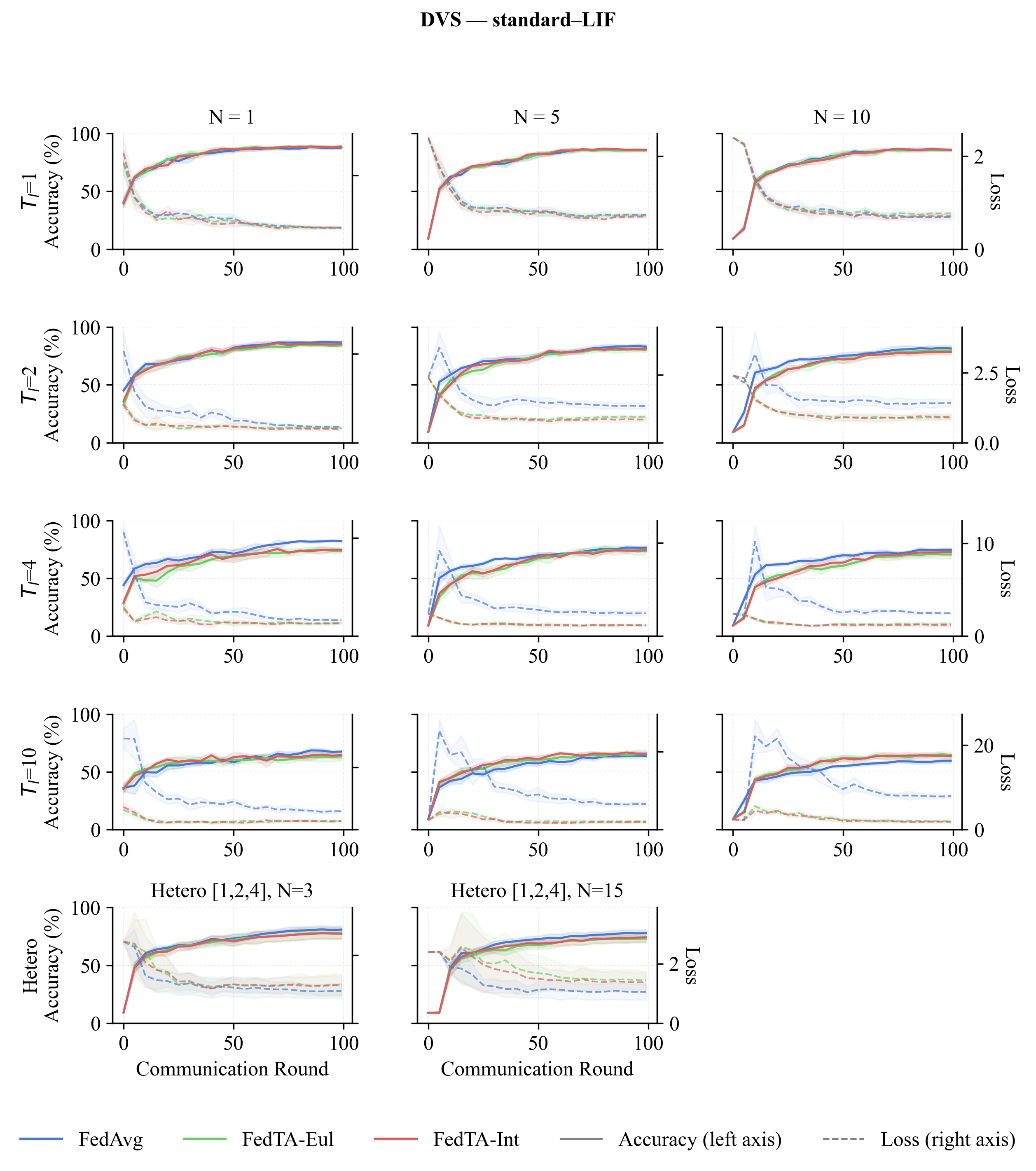}
\newpage
\includegraphics[width=1\linewidth]{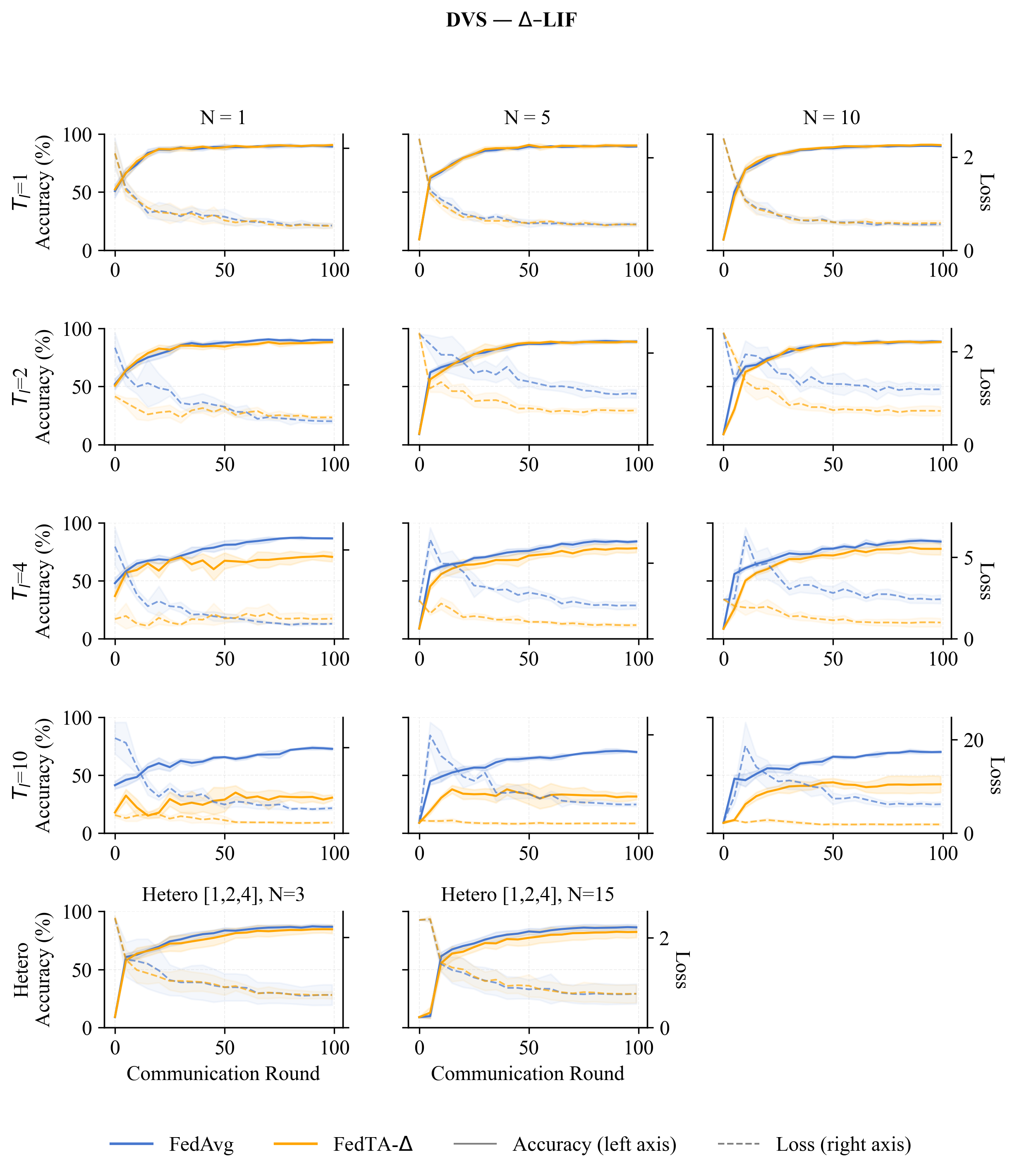}
\newpage
\includegraphics[width=1\linewidth]{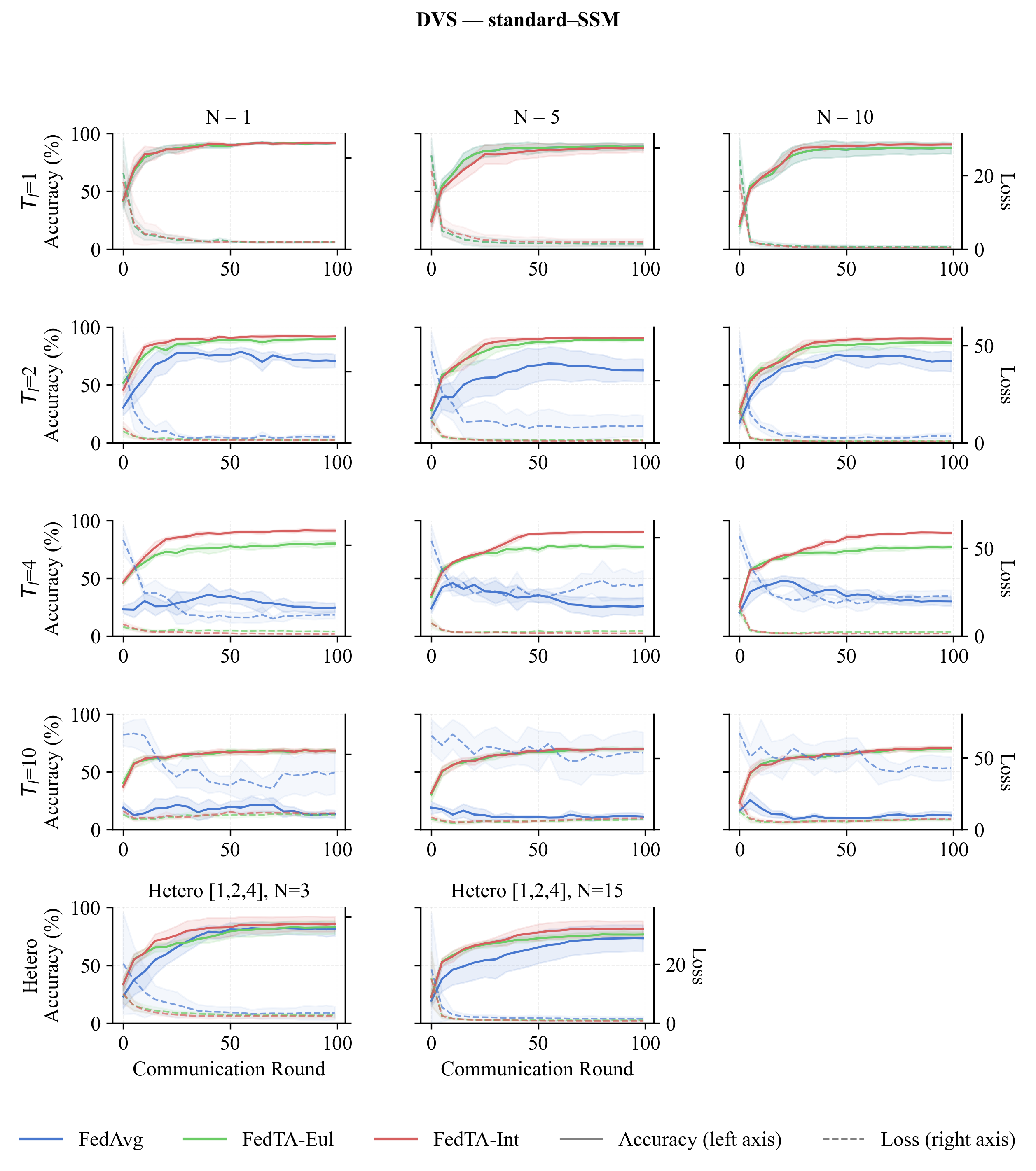}
\newpage
\includegraphics[width=1\linewidth]{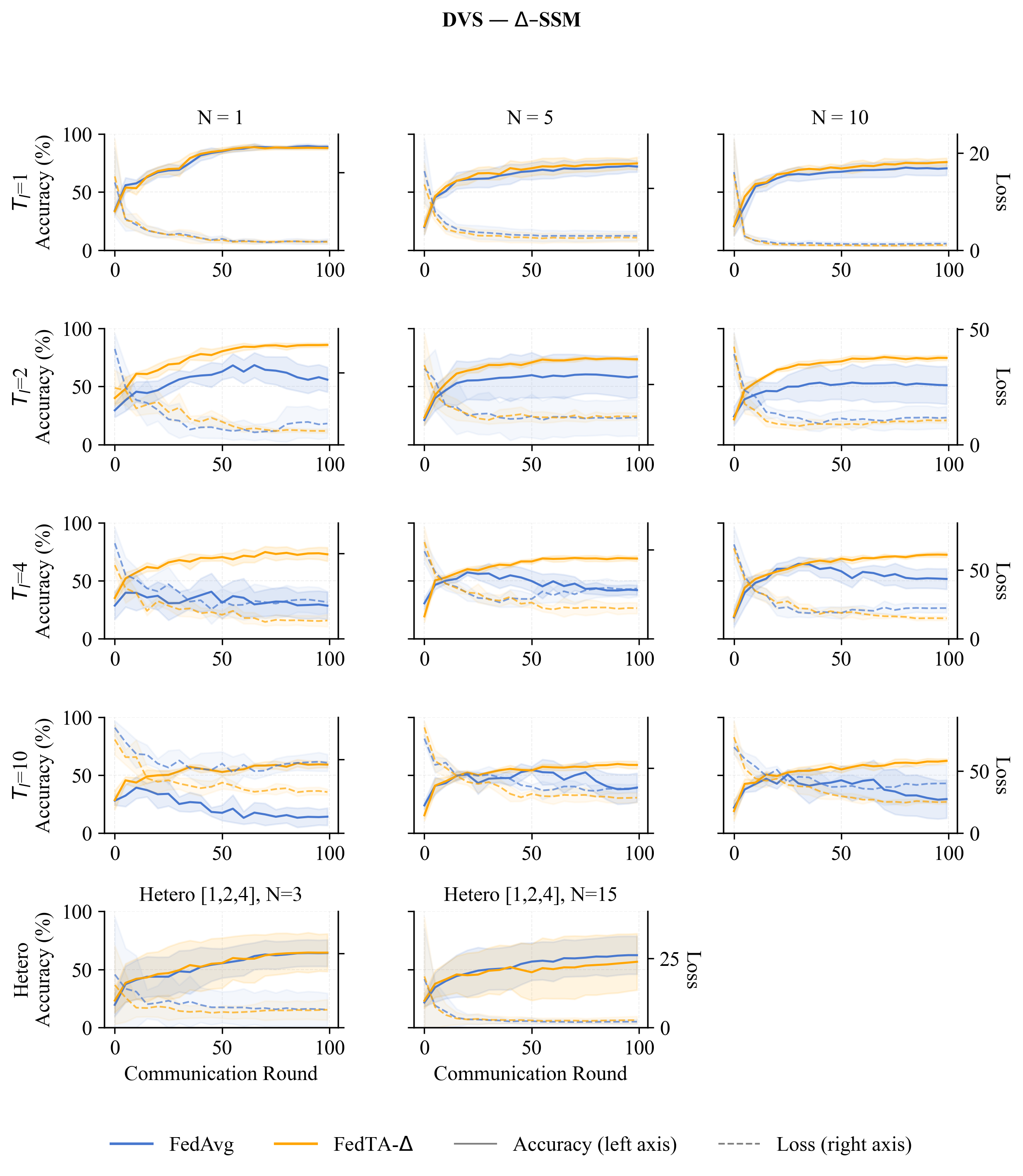}
\newpage

\end{document}